\documentclass[11pt]{article}

\usepackage{emnlp2021}
\usepackage{times}
\usepackage{latexsym}

\usepackage{url}
\usepackage{subcaption}
\usepackage{url}
\usepackage{graphicx}
\usepackage{enumitem}

\usepackage{comment}
\usepackage{tikz}
\tikzstyle{every picture}+=[remember picture]
\tikzstyle{na} = [shape=rectangle,inner sep=0pt,text depth=0pt]

  \usepackage{todonotes} 

\title{Empathic Conversations: A Multi-level Dataset of Contextualized Conversations}

 \author {
 Damilola Omitaomu$^1$\thanks{\ \ These authors equally contributed to this work.}, Shabnam Tafreshi$^2$\footnotemark[1], Tingting Liu$^{3,1}$, Sven Buechel$^4$, \\ {\bf Chris  Callison-Burch}$^1$,  {\bf Johannes Eichstaedt}$^5$, {\bf Lyle Ungar}$^1$, {\bf Jo\~ao Sedoc}$^6$\thanks{\ \  Corresponding author: jsedoc@stern.nyu.edu} \\
 $^1$University of Pennsylvania; $^2$University of Maryland:ARLIS; $^3$National Institute on Drug Abuse; \\$^4$Friedrich-Schiller-Universit\"at Jena; $^5$Stanford University; $^6$New York University \\
}

\date{}

\begin{document}
\maketitle
\begin{abstract}
Empathy is a cognitive and emotional reaction to an observed situation of others. Empathy has recently attracted interest because it has numerous applications in psychology and AI, but it is unclear how different forms of empathy (e.g., self-report vs counterpart other-report, concern vs. distress) interact with other affective phenomena or demographics like gender and age.
To better understand this, we created the {\it Empathic Conversations} dataset of annotated negative, empathy-eliciting dialogues in which pairs of participants converse about news articles. People differ in their perception of the empathy of others. These differences are associated with certain characteristics such as personality and demographics. Hence, we collected detailed characterization of the participants' traits, their self-reported empathetic response to news articles, their conversational partner other-report, and turn-by-turn third-party assessments of the level of self-disclosure, emotion, and empathy expressed. This dataset is the first to present empathy in multiple forms along with personal distress, emotion, personality characteristics, and person-level demographic information. We present baseline models for predicting some of these features from conversations. 
\end{abstract}
\section{Introduction}
\begin{figure}[t!]
    \centering
    \includegraphics[width=\linewidth]{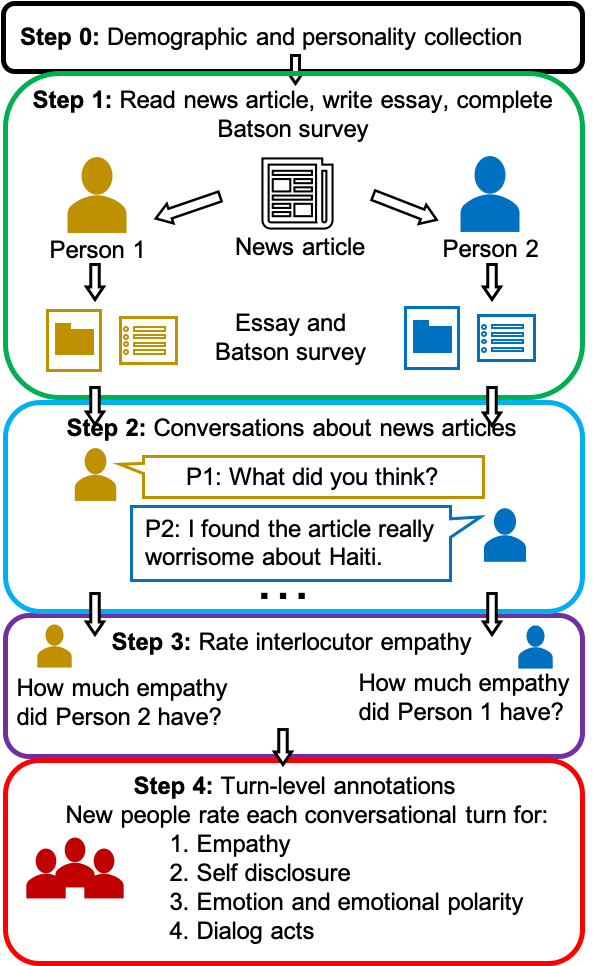}
    \caption{A schematic of data collection. First, (Step 0) we collect the demographic and personality information of participants using a survey, then (Step 1) each participant writes an essay about the news article and submits the Batson survey~\cite{batson1987distress} to assess the self-reported level of empathy and distress the person felt toward the article. Next, (Step 2) two participants converse about an article, and at the end of their conversation, (Step 3) they rate other-report counterpart perceived empathy. Finally, (Step 4) other annotators (third-person assessment) label the conversational turns for Empathy, Self-Disclosure, Emotion, and Emotion Polarity, and dialogue acts.}
    \label{fig:collection-overview}
    \vspace{-0.4cm}
\end{figure}
Humans are an irreducibly social species. The complex social environment requires us to quickly and accurately process cues we received and correspondingly generate reactions~\citep{preston2002empathy}.
Affective states (embodied feelings, short term emotions, and longer-term moods) are particularly potent contextual cues and elements of the human experience, as they substantially impact almost every phase of our cognitive functioning and social interactions, including attention, perception, memory, and behavioral reactions. 
Thus, when interacting with conversational agents, their ability to understand and interact in emotionally appropriate ways has become more important to facilitate a successful social interaction, as its the basis for humans to feel seen, understood, and generate trust. One important way to understand and translate others' emotional expression is empathy.

Designing conversational agents informed by and responsive to empathy is important to effectively communicate with users and thus emerging area of research. 
For example, empathy is critical for clinical applications of agents such as automated behavioral therapy agents for self-destructive and unhealthy behaviors \citep{fitzpatrick2017delivering}. 
Many studies show that effective gesturing of robots should include natural language understanding of empathy \cite{fung2015robots}. In the field of embodied and virtual agents, empathy is critical and actively studied~\citep{Mcquiggan2007,paiva2017empathy,yalcin2018computational}. 

However, agents in AI and more generally human-computer interaction, have yet to successfully implement this complex emotional-motivational state. This is in part due to the lack of training data sets in which empathy can be modelled in the full complexity with which humans encounter it.

In this paper, we present a novel dataset that includes dyadic (two person) text conversations of crowd workers about news articles, thus complementing the first-person statements of the Empathic Reactions data by \citet{buechel2018modeling}.\footnote{\url{https://github.com/wwbp/empathic_reactions}} 
To obtain a more comprehensive empathy rating, we assessed our conversation from three different angles, including self-report empathy assessed from two aspects (compassion and distress); other-report second-person perceived empathy of counterpart in the conversation; and third-person turn-level annotation of empathy, self-disclosure, emotion, and emotional polarity (Figure~\ref{fig:collection-overview}). Additionally, we added assessments of demographics and personality, which have been found to be correlated with empathy perception, to provide a rich psychological constructs.
Our \textit{Empathic Conversations} dataset\footnote{The dataset is available by request.} provides a rich psychologically founded dataset with detailed analysis.\footnote{Our research was performed as a registered protocol under IRB\# 826448.}

Our contributions are as follows:
\begin{enumerate}[topsep=0pt, partopsep=0pt,noitemsep]
    \item{We present an empathetic conversation dataset grounded in responses to news articles; this includes empathy and distress of each conversant along with their personality and demographic information.}
    \item {We build models to predict self-report empathy and distress; second-person perceived empathy; and turn-level Empathy, Emotion/Emotional Polarity, and Self-Disclosure.}
    \item {We show the usefulness of the different levels of annotation for predicting multiple levels of empathy.}
\end{enumerate}

The rest of this article is structured as follows: Section \ref{relwork} describes the related resources and modeling of empathy and distress; Section \ref{data} explains the dataset statistics, annotation process, and challenges; Section \ref{models} provides deep-learning models for empathy and distress; Section \ref{disc} describes the findings in the annotation process and the models we built for empathy and distress;  Section \ref{concfu} concludes this study and suggests future directions; Section \ref{ethics} describes the ethical factors we considered during the annotation process.

\section{Related Work}
\label{relwork}
\textbf{Empathy: }
To build and evaluate the empathic conversations between human and AI agents, how human produce and process empathy should be aware. While conceptualized slightly differently by different theorists, emotional empathy includes two different types of processes: process of responding to others’ distress, and the process of experiencing the feelings on behalf of the observed person~\citep{goubert2005facing,chen2018empathy}. The latter one was usually termed as `empathic concerns' and here we use interchangeably with `empathy' in our study~\cite{lin2012empathy}.
For the responses to distress, it involves the representations of self and other in generating the a similar (negative) affective reactions (e.g., increased personal distress evoked by others' distress); for the empathy, it is a feeling of concern, sympathy, warm, tenderness, and compassionate (e.g., feel others' feelings~\citep{buechel2018modeling,lin2012empathy}). To assess these two important underpinnings of empathy in human-AI interaction, we utilized a first-person survey, the Batson's Empathic Concern -- Personal Distress Scale \cite{batson1987distress} (as suggested in \citet{buechel2018modeling}).

Empathy has also been found to be related to personality. For example, agreeableness and conscientiousness have been found to be the most important and consistent predictors of empathy in a survey research conducted across four countries~\citep{melchers2016similar}. To better understand empathy in human-AI conversational interactions, we also included self-report personality traits of participants.

\textbf{Modeling text-based empathy: } Prior work on modeling text-based empathy focuses on the following objectives and definitions:
\citet{litvak2016social,fung2016towards} studied the empathetic concern which is to share others' emotions in conversations; \citet{xiao2015rate,xiao2016technology,gibson2016deep} modeled empathy based on the ability of a therapist to adapt to the emotions of their clients; and \citet{zhou2020condolences} quantified empathy in condolences in social media using \textit{appraisal theory}.

Several works analyze and model text-based empathy:
\citet{khanpour2017identifying} proposed a neural network (NN) model to predict empathy in health-related posts. \citet{buechel2018modeling} modeled empathy and distress in crowdworkers' written reactions to (empathy-evoking) 
news articles.
In a follow-up study, \citet{sedoc2020learning} developed a mixed-level feed-forward NN model that learned \textit{word-level} empathy and distress from the above text-level ratings, resulting in a lexicon of empathy and distress words. Recent work by \citet{hosseini2021takes} studied whether each person in an online health community seeks or wants to provide empathy. Empathy was also studied in the context of virtual and embodied agents \citep{yalcin2018computational,paiva2017empathy}.

\textbf{Dialogue Systems with Empathy Capabilities:} To date, most dialogue systems with empathy capabilities are trained on conversations grounded in personal situations in order for agents to learn to seem more engaging and empathetic.

The Empathetic Dialogues (ED) dataset \citep{rashkin-etal-2019-towards} consists of 25k personal conversations. Each dialogue is grounded in specific emotional situations where a speaker feels a certain emotion towards a circumstance and receives empathetic responses from the other speaker. ED has served as a foundational dataset for building more empathetic dialog models~\citep[e.g.,][]{lin-etal-2019-moel,majumder-etal-2020-mime,li-etal-2020-empdg,smith2020can,gao-etal-2021-improving-empathetic}. There have been several efforts to increase the size of ED with filtered found data. \citet{welivita-etal-2021-large} curated a `silver-standard' dataset Emotional Dialogues in OpenSubtitles (EDOS) of 1M dialogues annotated in semi-automated way into nine emotion classes. The persona-based
empathetic conversation (PEC) dataset is mined from Reddit comments resulting in 355k empathetic conversations with persona information~\citep{zhong-etal-2020-towards}.

Another set of work aimed to create a dataset that is motivated by psychological theory of empathy. \citet{sharma-etal-2020-computational} introduced the EmPathy In Text-based, asynchrOnous MEntal health conversations (EPITOME) dataset in an effort to detect empathy in textual conversations in a theoretically-grounded way. This dataset was used in subsequent work to rewrite counselor text in a more empathetic and effective manner~\citep{sharma2022human}.

Our dataset is akin to these datasets in that we capture empathy in conversations; however, we ground these conversations by news articles and capture multiple views and dimensions of empathy. Similar to the EPITOME dataset, our EC dataset is motivated by psychological theory. 

Several works are grounded in personal context to make agents more engaging for users \citep{li2016persona,vijayakumar2016diverse,mazare2018training}. They often use another emotion-annotated dataset is the DailyDialog (DD) dataset \citep{li2017dailydialog}. It consists of daily communications crawled from different websites used for English learners to practice English dialog in daily life. DD is annotated manually with emotion and the purpose of the communication. 
\section{Data Acquisition and Annotation Methods}
\label{data}
In our work, we study how a range of personality traits affect conversations, specifically empathetic conversations. Our dataset consists of 500 conversations between crowd workers chatting through a text interface about a selection of articles from \citet{buechel2018modeling}. We use the top 100 articles of negative events which elicited the most empathy and personal distress. We note that this study analyzes the negative empathy, in addition, we cannot measure whether our annotators' reactions were natural/wild or made up. They were aware that they would read negative events, but they did not know the event topic. Hence,  wild or made up reactions are possible. We quantify the empathy in conversations that are grounded in these news articles. In the following subsections, we explain the collection procedure (Figure~\ref{fig:collection-overview}) for our {\it Empathic Conversations} dataset. ~\autoref{tab:data-stats} details the total number of annotations. 

\subsection{Data Collection and Annotation Methodology}
\paragraph{Step 0: Recruitment and Demographic / Personality Information}
The acquisition of participants was set up as a crowdsourcing task on MTurk.com pointing to a Qualtrics questionnaire. 
We collected demographic and personality information, including the widely used Big Five (OCEAN) personality traits \cite{john1999big}\footnote{We use the Ten Item Personality Inventory \cite[TIPI;][]{gosling2003very}.} and Interpersonal Reactivity Index \cite[IRI;][]{davis1980interpersonal}. 
After the participants filled out background information dealing with demographics and personality, they then read a random selection of five news articles selected from the topmost 100 empathic articles. After reading each news article, each participant was then asked to rate their level of empathy and distress before summarizing their thoughts and feelings about the article. Each message was reviewed manually to filter out the responses that deviated from the task. Of the 110 workers that did the HIT, 92 workers performed it appropriately\footnote{Completed the survey and wrote proper essays and did not randomly fill out surveys. This is the same criteria as \citet{buechel2018modeling}.} and were offered a qualification to participate in the second crowdsourcing task on MTurk.

\paragraph{Step 1: Reading News Articles}
Before every conversation, each pair of crowd workers were asked to read a news article from the topmost empathy eliciting articles of negative events from \citet{buechel2018modeling}.\footnote{Available at \url{https://drive.google.com/file/d/1A-7XiLxqOiibZtyDzTkHejsCtnt55atZ/view}.} 
The annotation process was set up as a task on Amazon Mechanical Turk. Workers were grouped in pairs. 
For each of the 100 articles we collected 5 conversations.

\paragraph{Step 2: Essay and Batson Survey}
After reading the article, crowd workers were asked to write an essay (limit 300-800 characters) just as in \citet{buechel2018modeling}. During this phase, participants rated their empathy and distress level using the Batson scale and summarized their thoughts and feelings through a Qualtrics survey. 

\paragraph{Step 3: The Conversation}
Next, participants were asked to express their empathy towards the article in an online text conversation with each other (Figure \ref{fig:collection-overview}). We collected dialogues using the ParlAI platform to interact with Amazon Mechanical Turk participants, who were paired together to have a $\geq$ 15 turn conversation about the article. We provide a snapshot of an example news article and crowd workers' conversation (Figures \ref{fig:conv_article} and \ref{fig:convo} in Appendix \ref{sec:supplemental}). Workers were paid \$1 per HIT completed.

\paragraph{Step 4: Other-report  Empathy Rating} 
Further, we obtained \textit{other-report empathy} by asking each annotator to rate the level and nature of empathy they perceived in each other (second-person) on a scale of 1-7. Other-report counterparty empathy is collected after the conversation has been completed.

\paragraph{Step 5: Turn Level Annotations of Conversations}
\label{dataturn}
Next, we collected third-person annotations of every conversational utterance from crowd workers. Turn level annotations on the collected conversations were conducted using Amazon Mechanical Turk and the same workers that participated in the collection of the conversations. The workers were asked to analyze the following aspects of each turn: Empathy, Emotion, Emotional Polarity, and  Self-Disclosure. In Appendix \ref{sec:supplemental}, Figure \ref{fig:turnempathy} illustrates a snapshot of the interface we presented to the workers. In each HIT, workers were asked to annotate only one aspect (e.g., only Empathy) for an entire conversation. Workers were paid \$0.25 for each HIT completed with a \$0.75 bonus for good work.

Finally, 54 conversations (1,400 out of 5821 individual turns) were labeled by dialogue act, by 1 annotator for each turn, using the annotation manual from \citet{Jurafsky1997SwitchboardSS}. These were sufficiently difficult to crowdsource 
such that we had research assistants annotate the turns. After training, 92\% agreement was reached on annotations.\footnote{The remaining 8\% were ambiguous.} Subsequently, the conversational turns were coded without redundancy; however, any unclear situations were discussed. Many turns for an individual contained multiple dialog acts. As expected, 30.5\% of utterances contained opinion statements, 17.4\% included agreement, and 7.3\% were factual statements. 

\subsection{Task Implementation}
\label{taskimp}

The main issue during this stage was the wait time for workers to be matched with other workers. We implemented several solutions to improve wait time efficiency. First, we sent out specific times when HITs would be posted and then noted which 10 minute intervals workers should join the HIT. If the wait time exceeded the 10 minutes to match with someone, we compensated the workers for the wait times. Despite all these efforts to efficiently use worker's time, there were still cases where the wait time exceeded 10 minutes, hence we had to dismiss those workers. Second, we implemented a sign-up sheet that would allow us to specifically post HITs when most workers stated they would be available. Only 25 / 92 workers filled out this sign-up sheet and there was a direct correlation between whether a worker joined outside the specified times and whether they completed the sheet. 
Additionally, based on feedback from early runs, we reduced the minimum number of required conversational turns by workers to 15 turns. We offered bonuses to incentivize longer turns: 15-20 turns a \$1.50 bonus and 20+ turns a \$2 bonus.

The main challenge with collecting \textit{Turn Level Annotations} was to find the best description for the categories to offer a more uniform understanding, since workers had different viewpoints on the meaning of Empathy and Self-Disclosure when annotating. 
The workers who completed the Turn Level Annotation were the same as the ones who participated in the initial conversation collection.\footnote{$<$ 3\% of turn-level annotations are by conversants in the conversation and can easily be removed without effect, since turn-level annotations are 3-way redundantly annotated.}
\subsection{Dataset Statistics}
\label{dataana}

\begin{table}[ht!]
    \centering \scalebox{0.9}{
    \begin{tabular}{cc}
    \hline
         \textbf{The Hub}&  \\ \hline
         New Articles & 100 \\
         Conversations & 500  \\ \hline
         \textbf{The Spokes} & \\ \hline
         Person-level  &  \\
         Personality \& Demographic Info & 79 \\ \hline
         Post-article /pre-conversation & \\Essays - Batson (Empathy/Distress) & 1,000 \\ \hline
         Post-conversation Perceived Empathy & 1,000 \\ \hline

         Turn level Annotations& \\
         (Empathy, Self-Disclosure, Emotion) & 5,821\\ 
         Turn - level dialog Acts & 1,400 \\ \hline

    \end{tabular}}
    \caption{Corpus statistics detailing the number of annotations.}
    \label{tab:data-stats}
\end{table}

\begin{figure}[t!]
    \centering
    \includegraphics[width=\linewidth]{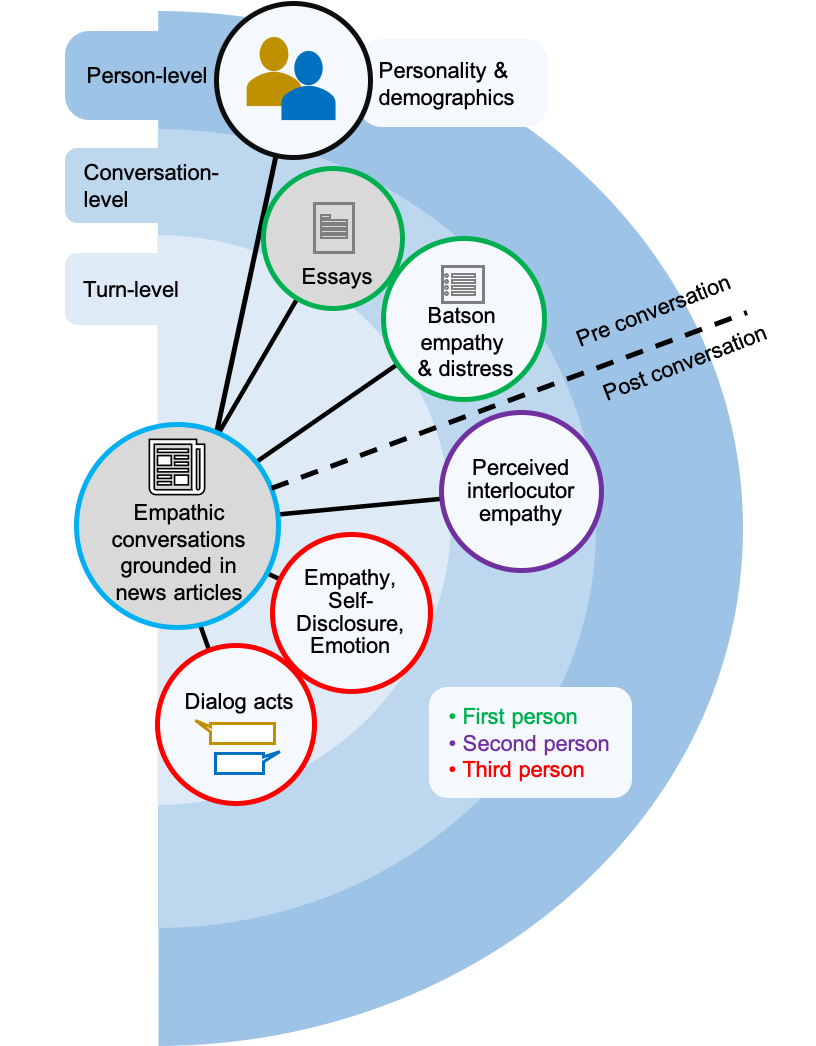}
    \caption{\textit{Empathic Conversation} dataset combines person-, conversation-, and turn-level information. Furthermore, there are multiple first-, second-, and third-person perspectives. Gray shading indicates text components of the dataset.}
    \label{fig:ec-multi-level}
\end{figure}

In this subsection, we detail the basic statistics of the collected dataset (\autoref{tab:data-stats} and \autoref{fig:ec-multi-level}). This dataset consists of 500 conversations that each took an average of 30 minutes (min=12, max=65) to be completed. 75 US-based Mechanical Turk workers participated in this study. Each worker was paired by time priority (i.e. the first 2 workers to accept the HIT were paired). As a result, the same workers were paired together in multiple conversations. The 75 workers completed an average of 13 conversations (min=1, max=88). The highest frequency of separate conversations that two workers held together was 11, within the 251 distinct pairs of worker conversations; 4 other worker pairs had 10 conversations together. 
The average age of the workers is 35 years old (min=19, max=62) with an average income of 58,000 (min=5,000, max=165,000). 
Furthermore, gender was more male-dominated with 44 male workers and 31 female workers. 
The workers come from various educational and ethnic backgrounds with almost half having a 4 year bachelor's degree and more than half being White (for further details see Tables \ref{table:education} and \ref{table:race} in \autoref{sec:demo} of the datasheet). 

\begin{figure}[tbh]
    \centering
    \includegraphics[width=0.4\textwidth]{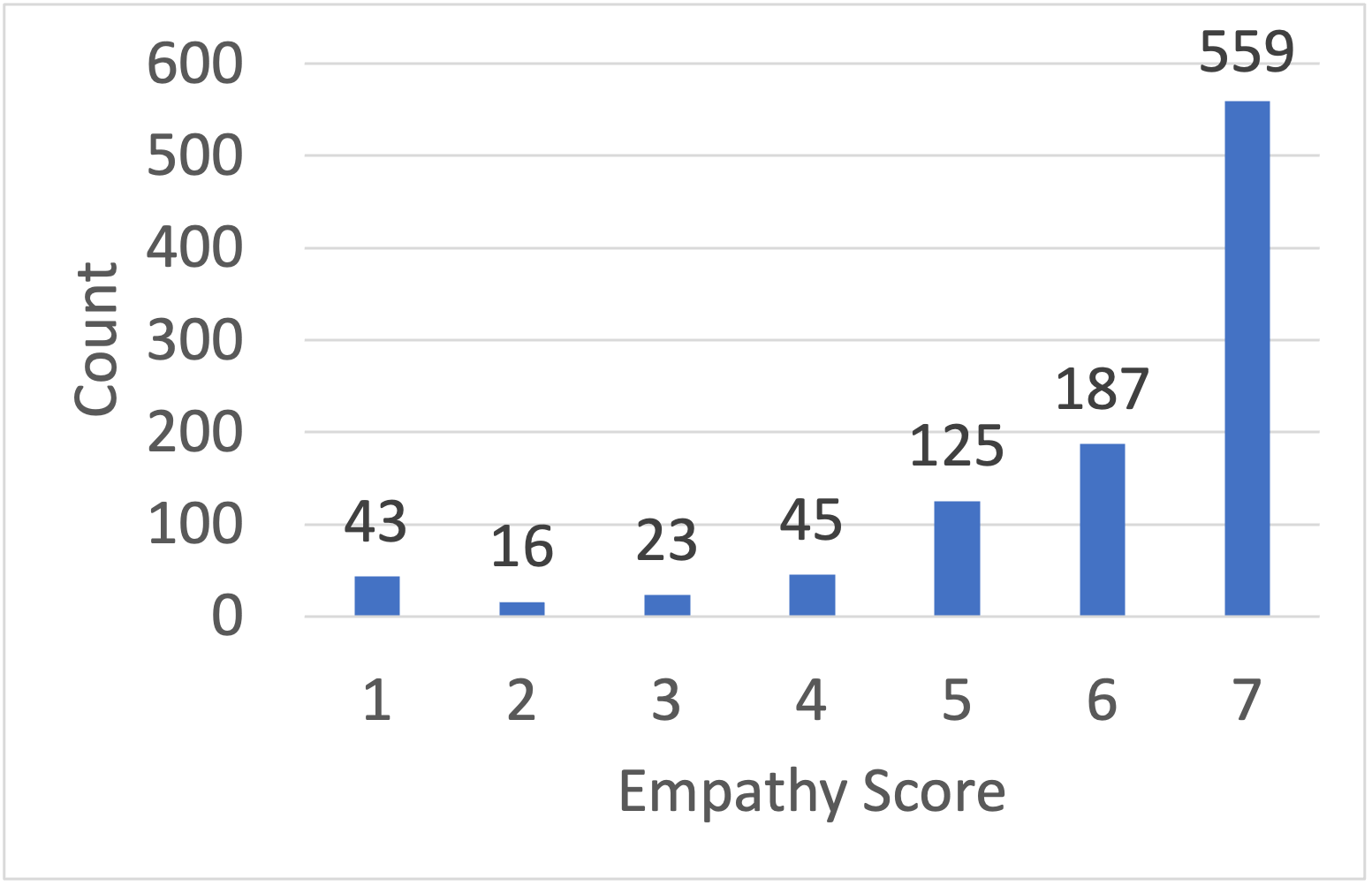}
    \caption{Perceived empathy level counts.}
    \label{fig:perceived_empathy_hist}
\end{figure}

Data analysis is based on the qualification survey data. Each worker had to take the survey before entering into any conversation. The \textit{perceived counterparty empathy score} was rated at the end of the conversation. Since the workers were told to discuss their thoughts and feelings about the article, they had a lot of freedom to steer the discussion. Each conversation was manually examined to ensure that the workers were having a valid conversation. We noticed that the conversations are of high quality with the workers drawing in personal relations to the topic. One concern was that workers would focus on the factual information within the articles instead of expressing their opinions and feelings. Interestingly this did not occur. Another concern was that the conversation would be controlled by the first speaker and turn into more of an interrogation, resulting in ``plain'' responses from the ``interrogated.'' We did not observe this in the majority of conversations;  both workers contributed equally to the substance of the conversation; however, we noticed instances in which the conversations did not have much substance.
Nonetheless, the distribution of empathy scores among the workers was clustered toward the higher end (\autoref{fig:perceived_empathy_hist}). 

\paragraph{Annotator Agreement}

The overall inter-annotator Agreement (IAA) measured using Krippendorff's alpha~\citep{hayes2007answering} was 0.492. 
Table \ref{tab:turn-level-iaa-by-type} shows the breakdown of IAA by Empathy, Emotion, Emotional Polarity, and Self-Disclosure. These alpha statistics are calculated by micro-averaging per conversation, and thus, none of the individual annotations are more than the overall Krippendorff's alpha. Low agreement is normal for fine-grained emotion(-like) annotation problems because this is a subjective task \cite{de-bruyne-etal-2020-emotional,troiano-etal-2021-emotion,demszky-etal-2020-goemotions,davani2021dealing}; however, when using mean absolute deviation filtering \cite{zhao-etal-2020-designing} and macro Krippendorff's alpha computation, all aspect alpha values are above 0.7.\footnote{We believe this is common ``alpha hacking'' and leave it up to the users of the dataset.}
Empathy has the lowest IAA which indicates the variety of this phenomenon among different personalities ({\bf note again that micro-average alphas are expected to be significantly lower}). Emotional Polarity and emotion have the highest IAA. This is expected because all articles are about negative events. Self-Disclosure has an average IAA among all these quantities.\footnote{See the \autoref{sec:turn-iaa-breakdown} for further discussion.} 
\begin{table}[ht!]
\centering
\begin{tabular}{lc} 
 \textbf{Category} & \textbf{Average IAA} \\ \hline
  Empathy & 0.274 \\ \hline
 Emotion & 0.389 \\ \hline
 Self-Disclosure & 0.335 \\ \hline
 Emotional Polarity & 0.442 \\ \hline
Overall & 0.492 \\ \hline
\end{tabular}
\caption{Turn level annotation inter-annotator agreement (IAA) using Krippendorff's alpha by category.}
\label{tab:turn-level-iaa-by-type}
\end{table}

\subsection{Multi-level Correlations Analysis}

One of the main contributions of the empathic conversations dataset is that we can look at interaction effects between various empathy measures, emotion, demographics, and personality. For all correlation statistical significance testing, we use a p-value $<0.05$ calculate via bootstrap analysis with multiple hypothesis testing correction.

First, we asked {\bf how are the multiple views of empathy correlated?} We found an asymmetry in that Person 1 (the initiator of the conversation) had a lower correlation between their actual empathy and the perceived empathy (Pearson correlation of 0.125) whereas Person 2 had a correlation of 0.35. There is a similar relationship for the correlation between perceived empathy and first-person distress where Person 1 distress is correlated to Person 2 perceived empathy of Person 1 at 0.125 whereas the other way is 0.262. This low correlation is consistent with prior work showing that people do not know how they appear~\cite{sun2019people}.
We found a weak correlation (0.13) between turn-level Empathy and perceived empathy; however, there is no significant correlation between turn-level Empathy and first-person empathy.

Next, we asked {\bf what is the relationship between demographic and empathy levels?}
We found that women showed higher distress than men; however, the empathy level was not statistically significantly different. Similarly, there was no significant difference in turn-level aspects. The perceived empathy in women was higher than men ($0.19$). We found that older participants ($>35$) showed lower first-person empathy and distress; however, they had higher turn-level Emotion and Self-Disclosure. 
Higher income individuals ($>58,000$) showed the same pattern as age with only one difference which is that their distress was much lower ($-1.08$). 
Higher education was correlated with higher first- and second-person empathy and distress, but no significant difference in turn-level aspects.
For this analysis, we found that there were insufficient racial set sizes to find statistically significant results. The only statistically significant other-report empathy difference was due to education level.

\begin{table*}[t]
\resizebox{\textwidth}{!}{%
\begin{tabular}{lccccccc}
\textbf{Psych Construct /   Observation}          & \multicolumn{1}{l}{\textbf{\begin{tabular}[c]{@{}l@{}}Self-report \\ Empathic Concern\end{tabular}}} & \multicolumn{1}{l}{\textbf{\begin{tabular}[c]{@{}l@{}}Self-report \\ Personal Distress\end{tabular}}} & \multicolumn{1}{l}{\textbf{\begin{tabular}[c]{@{}l@{}}Other-report \\ Perceived Empathy\end{tabular}}} & \multicolumn{1}{l}{\textbf{\begin{tabular}[c]{@{}l@{}}Turn-level \\ Empathy\end{tabular}}} & \multicolumn{1}{l}{\textbf{\begin{tabular}[c]{@{}l@{}}Turn-level \\ Emotion\end{tabular}}} & \multicolumn{1}{l}{\textbf{\begin{tabular}[c]{@{}l@{}}Turn-level \\ Emotional Polarity\end{tabular}}} & \multicolumn{1}{l}{\textbf{\begin{tabular}[c]{@{}l@{}}Turn-level \\ Self-disclosure\end{tabular}}} \\
\multicolumn{1}{c}{\textbf{Big 5}} & \multicolumn{1}{l}{}                                                                                 & \multicolumn{1}{l}{}                                                                                  & \multicolumn{1}{l}{}                                                                                   & \multicolumn{1}{l}{}                                                                       & \multicolumn{1}{l}{}                                                                       & \multicolumn{1}{l}{}                                                                                  & \multicolumn{1}{l}{}                                                                               \\
{\bf Openness}                                          & $\uparrow$                                                                                           & $\uparrow$                                                                                            & $\uparrow$                                                                                             & $\downarrow$                                                                               & $\downarrow$                                                                               & $\downarrow$                                                                                          & $\downarrow$                                                                                       \\
\textbf{Conscientiousness}                                   & $\uparrow$                                                                                           &                                                                                                       & $\uparrow$                                                                                             & $\uparrow$                                                                                 &                                                                                            &                                                                                                       &                                                                                                    \\
\textbf{Extroversion}                                      & $\downarrow$                                                                                         & $\downarrow$                                                                                          & $\uparrow$$\uparrow$                                                                                   &                                                                                            & $\uparrow$                                                                                 & $\uparrow$                                                                                            & $\uparrow$$\uparrow$                                                                               \\
\textbf{Agreeableness}                                     & $\uparrow$                                                                                           &                                                                                                       & $\uparrow\uparrow$                                                                                   &                                                                                            &                                                                                            &                                                                                                       &                                                                                                    \\
\textbf{Stability}                                         & $\uparrow$                                                                                           & $\downarrow$                                                                                          &                                                                                                        &                                                                                            &                                                                                            &                                                                                                       &                                                                                                    \\
\multicolumn{1}{c}{\textbf{IRI}}   &                                                                                                      &                                                                                                       &                                                                                                        &                                                                                            &                                                                                            &                                                                                                       &                                                                                                    \\
\textbf{Perspective-taking}                                & $\uparrow$                                                                                           &                                                                                                       & $\uparrow$                                                                                             &                                                                                            &                                                                                            &                                                                                                       &                                                                                                    \\
\textbf{Distress}                                          &                                                                                                      & $\uparrow$                                                                                            & $\downarrow$                                                                                           &                                                                                            &                                                                                            &                                                                                                       &                                                                                                    \\
\textbf{Fantasy}                                           & $\uparrow$                                                                                           & $\uparrow$                                                                                            &                                                                                                        &                                                                                            &                                                                                            &                                                                                                       &                                                                                                    \\
\textbf{Empathic concern}                                  & $\uparrow$                                                                                           & $\uparrow$                                                                                            & $\uparrow$                                                                                             &                                                                                            &                                                                                            &                                                                                                       &                                                                                                   
\end{tabular}%
}
\caption{Relationship between personality traits and various empathy perspectives.}
\label{tab:multiref-analysis}
\end{table*}

{\bf What is the correlation between empathy levels and personality?} 
Extroversion and Agreeableness are the most significant personality traits to increase perceived empathy (\autoref{tab:multiref-analysis}). In the case of Extroversion, we observe a disconnect between self-report and other-report perceived empathy. 
As expected, higher Openness leads to higher self-report empathy and distress and higher other-report perceived empathy; however, {\bf lower} turn-level Emotion, Empathy, and Self-Disclosure. 
Higher Conscientiousness follows a similar pattern of increased self- and other-report empathy; however, only the turn-level Empathy is higher.
Surprisingly, Emotional Stability seems to have no discernible effect. We conclude that to influence other-report empathy only two personality traits matter.

Higher perspective-taking correlates with higher self- and other-report empathy. 
Higher IRI distress correlates to higher self-report distress and lower perceived empathy. 
Higher IRI fantasy corresponds to higher self- and other-report empathy and distress. 
 Although most results are consistent with expectations, they help validate the annotation and guide possible personalities for a conversational agent.

Finally, we examined the impact of simple textual features on other-report perceived empathy. We provide statistics such as correlations of empathy score with respect to word count (total words they used), word diversity, word frequency, pronoun usage, and determiner usage. The average word count among all the conversations is 249  (min=52, max=1010), which has a weak positive correlation of 0.074. \textit{Word Frequency} has a correlation of 0.291. Pronoun usage has a correlation of 0.468. Determiner has a correlation of -0.051. Thus the pronoun usage has the highest correlation with the perceived empathy scores given by the workers. In conversations, whenever the worker was able to relate to the topic of the article personally, the worker was able to have a higher empathy. This finding is consistent with previous work on language emotionality and word statistics \citep{tausczik2010psychological}.

\section{Modeling Empathy and Distress}
\label{models}
We developed the baseline models to show the impact of our dataset in predicting empathy, distress, and perceived empathy in text. First, we modeled \textit{empathy} in the turn-levels in each conversation. Second, we modeled the perceived counterparty empathy in conversation level. Lastly, we modeled self-report Batson scale empathy and distress.\footnote{The reason for choices in pre-trained LM (i.e., RoBERTa) and NN architecture like bi-rnn is our observations in several successful results that are obtained in prior text classification studies such as sentiment and emotion classification.}

\subsection{Modeling Empathy}
\label{modempathydistress}
\paragraph{Task setup} We implemented models to predict Empathy, Emotion, Emotion polarity, and Self-Disclosure for each turn in the turn-level conversations. We developed two models: we trained a Gated Bi-RNN with attention layer (Bi-rnn-Att) and we \textit{fine-tuned} the RoBERTa-base pretrained language model with our dataset. We use \textit{fasttext} model (300-dimensions) in \textit{Bi-rnn-Att} to represent the context features for each turn, then we concatenate these context features with numerical values of Empathy, Emotion, Emotion Polarity, and Self-Disclosure. For example these numerical values to model empathy are: Emotion, Emotion Polarity, and Self-Disclosure. We refer to these numerical values \textit{features} in the entire of this section. To evaluate our dataset with transformer models we fine-tuned RoBERTa-base pretrained LM, 
we used [CLS] token for \textit{fine-tuning}.
There are a total of 5821 individual turns in all turn-level conversations. We split the data randomly into 70\%/15\%/15\% for train/dev/test, respectively. Training and tuning conditions are provided in the Appendix. 

\paragraph{Results} Table \ref{table:8}) demonstrates the results. 
Best result is obtained by fine-tuning the RoBERTa-base model. We can observe that when \textit{features} are added to each model in Bi-rnn-att condition it improved the results for each metric, and the results of Bi-rnn-att-features are comparable with RoBERTa-base model for each metric. Particularly we can observe the impact of these numeric \textit{features} for modeling \textit{Emotion}, \textit{Emotion Polarity} and \textit{self-disclosure}. One observation is the significant impact of \textit{features} in modeling \textit{Emotion, Emotion Polarity, and Self-disclosure}. In all these models numerical value of \textit{Empathy} is among the \textit{features} that is concatenated with the context features. As future study it is worth to analyze the impact of this feature in more isolated setup.

\begin{table}[ht!]
\begin{center}
\scalebox{0.7}{
\begin{tabular}{ccccc}
      \hline
      \textbf{Conditions} & \textbf{Empathy} & \textbf{Emo} & \textbf{Emo-pol} & \textbf{Self-dis} \\
      \hline
      Bi-rnn-att           & 0.575 & 0.369 & 0.339 & 0.370\\
      Bi-rnn-att-features  & 0.613 & 0.706 & 0.609 & 0.638\\
      RoBERTa-base         & \textbf{0.771} & \textbf{0.814} & \textbf{0.812} & \textbf{0.769}\\
      \hline
\end{tabular}}
\caption{Model performance for predicting turn-level Empathy, Emotion, Emotion Polarity, and Self-Disclosure in Pearson \textit{r} for each turn in turn-level conversations. ``Features'' is the numeric value of each of these quantities concatenated to the encoded sentences (e.g., for empathy we concatenated the numeric values of Emotion, Emotion Polarity, and Self-Disclosure as features). }
\label{table:8}
\end{center}
\end{table}
\subsection{Modeling \textit{Perceived Counterparty Empathy}}
\label{sec:perceived-empathy-model}
\paragraph{Task setup}  We used the same NN architecture described in subsection \ref{modempathydistress}. All models were trained to predict \textit{Perceived Counterparty Empathy} for the entire conversation. 
\paragraph{Results} We found that the Perceived Counterparty Empathy of person 2 is more predictable than person 1 (\autoref{table:9}).  Table \ref{table:9} shows the results that achieved when person 1 scored the \textit{Perceived Counterparty Empathy} from person 2 and vise versa.\\
\begin{table}[ht!]
\begin{center}
\scalebox{0.7}{
\begin{tabular}{cccc}
      \hline
      \textbf{Model} & \textbf{(Person 1)} &  \textbf{(Person 2)}\\
      \hline
      Bi-rnn-Att & 0.268 & 0.115 \\
      \hline
\end{tabular}}
\caption{Model performance for predicting other-report perceived empathy in Pearson \textit{r} for two people scoring at the end of the conversation.}
\label{table:9}
\end{center}
\end{table}
\subsection{Modeling \textit{Empathy} \& \textit{Distress} in Essays}
\paragraph{Task setup} We \textit{fine-tuned} RoBERTa-base with essays that both person 1 and 2 wrote after reading the article. For this series of experiments, we split the data into 60\%/10\%/30\% for train/dev/test, respectively. The reason for this choice of split is this dataset is smaller relative to conversation dataset, hence, bigger split for test set was necessary to have a reasonable numebr of datapoints in test set. 

\paragraph{Results} Table \ref{table:10} shows the models' performance in predicting \textit{empathy} and \textit{distress} based on the context of the essays person 1 and 2 wrote after reading the article.\\
\begin{table}[ht!]
\begin{center}
\scalebox{0.7}{
\begin{tabular}{cccc}
      \hline
      \textbf{Model} & \textbf{Empathy} & \textbf{Distress} & \textbf{Mean} \\
      \hline
      RoBERTa-base & 0.560 & 0.665 & 0.612 \\
      \hline
\end{tabular}}
\caption{Model performance for predicting \textit{empathy} and \textit{distress} in Pearson \textit{r} for essays written by annotators after reading the articles.}
\label{table:10}
\end{center}
\end{table}

\section{Discussion}
\label{disc}

We demonstrate the modeling of Emotion, Emotional Polarity, Self-Disclosure, Empathy and Distress. Further, personality and demographic variables are important when personalized empathy is the goal of an application.

The empirical results we obtained from Deep NN models suggest that our dataset is sufficient for modeling empathy. Further, our results indicate that emotion, polarity, and self-disclosure improve the modeling of empathy. Despite the complexity of empathy and the difficulties some people may have being empathetic or expressing feelings of empathy, our results in modeling empathy are significant. 

\textit{Perceived Counterparty Empathy} is also a significant feature of this dataset. This feature can be used to measure quantities in downstream tasks or applications, especially, ones that measure the amount of empathy. It is interesting that the model performs much better on Person 1, who initiates the conversation, than Person 2 (Table~\ref{table:9}). The reason for such a result is that Person 1 is on average less empathetic than Person 2. This is likely to be because  the first person to read the article and complete the survey starts the conversation. 

\section{Applications \& Future Work}
\label{apps-future}

The many facets of the {\it Empathic Conversation} dataset make it useful for number of applications and future research. Here are a few of these:

\begin{itemize}[topsep=0pt, partopsep=0pt,noitemsep]
    \item {Developing conversational agents that are perceived as empathetic will lead to more successful interactions; for example, expressing empathy is crucial for behavioral therapy agents, and more generally, understanding empathy in a relational framework is critical~\citep{vandijke2020towards}.}
      \item {Enabling personality traits in conversational agents is promising, as
    conversational agents displaying certain personality traits are more likely to be perceived as empathetic, and hence more successful in their interactions ~\citep{costa2014associations}. Agreeableness, and perhaps also openness and sociability are likely to be beneficial~\citep{hojat2005empathy}. }
    \item {Training empathic agents to make use of context (here, both the news articles and the demographics and personalities of the people conversing) may enable agents to express empathy in more natural and convincing ways.}
    \item{There are noticeable differences between how people experience their own empathy, that of the people that they are conversing with, and conversations that other people are having.  Teasing apart the linguistic markers of these different perceptions is key to understanding them.}

\end{itemize}
This dataset will facilitate future work including a better understanding of how demographics and personality correlate with empathy and distress in conversations.

\section{Conclusion}
\label{concfu}
Empathy and distress are crucial components in human-computer interaction and important traits of conversational AI agents. In this work, we provide an empathic conversations dataset grounded in news articles, where 2 people express their empathy about the plight of a third party around a shared real-world topic of conversation. Such data provides a new perspective on empathic conversations.Further, at the end of the conversation, each worker rated the amount of empathy they perceived from their counterpart. This is the first attempt to collect empathic conversations with these characteristics. Further, we collected demographic variables and personality of the authors of the conversations and we analyzed the correlation and importance of personality with empathy and distress. 
This is particularly important for personalized empathic AI agents and other applications in human-computer interactions. 

\section{Ethical Considerations}
\label{ethics}
There are two main ethical concerns regarding our dataset: the first is the possibility of disclosing private information\footnote{We follow HIPPA guidelines \url{https://www.hhs.gov/hipaa/for-professionals/privacy/special-topics/de-identification/index.html}.} and the second is the possibility of misuse of the data for manipulation of others (e.g., a malicious influence bot).
Regarding data privacy, all identifiable information was removed from the released dataset including Amazon Mechanical Turk IDs. All IP information, as well as the time of collection, have been removed. 
We also read all of the conversations to ensure that no personally identifiable information was present. This is especially important given the high level of self-disclosure in our dataset.

Our research was performed as a registered protocol under IRB\# 826448. Participants were told that these conversations and their demographic, personality surveys, and essays would be used for research purposes and distributed externally for research purposes.
\begin{quote}
    Your response to the survey will be used publicly for academic research. Your personally identifiable information will be stored on our secure server and never shared with third parties. We will use your data as part of publicly available research dataset, and will only report de-identified information, so no one can identify you as an individual person.
\end{quote}

Nonetheless, to mitigate this risk we make this data accessible only to research communities and we emphasize the risk involved in using this dataset. 
The corpus will be available under an Academic License (free of cost). Researchers will download a pdf and will supply their intended use. The process will require the institution to sign the agreement. The license will be a standard FDP data transfer form. It should be easy for institutions to obtain a signature.
 
The training models could potentially be misused to predict empathy, demographics, and personality for persuasion, given that the Empathic Conversations dataset is sufficiently rich. By distributing the dataset with clear requirements this risk will be mitigated. 

\bibliography{anthology,lit}
\bibliographystyle{acl_natbib}

\clearpage

\appendix
 
\section{Turn-Level Annotations}
\label{sec:supplemental}
\begin{figure}[ht!]
    \centering
    \includegraphics[width=8cm]{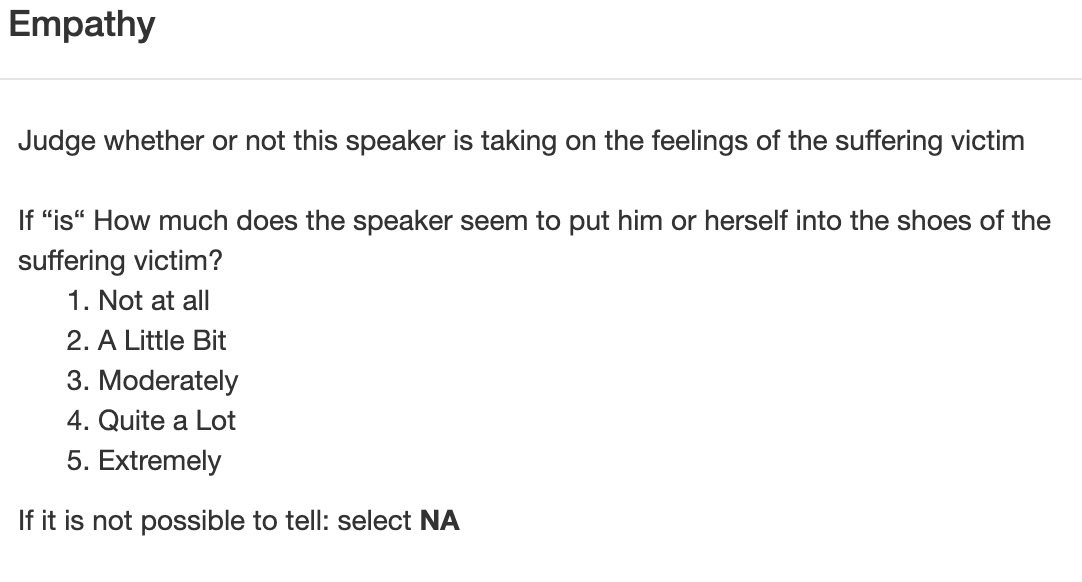}
    \includegraphics[width=8cm]{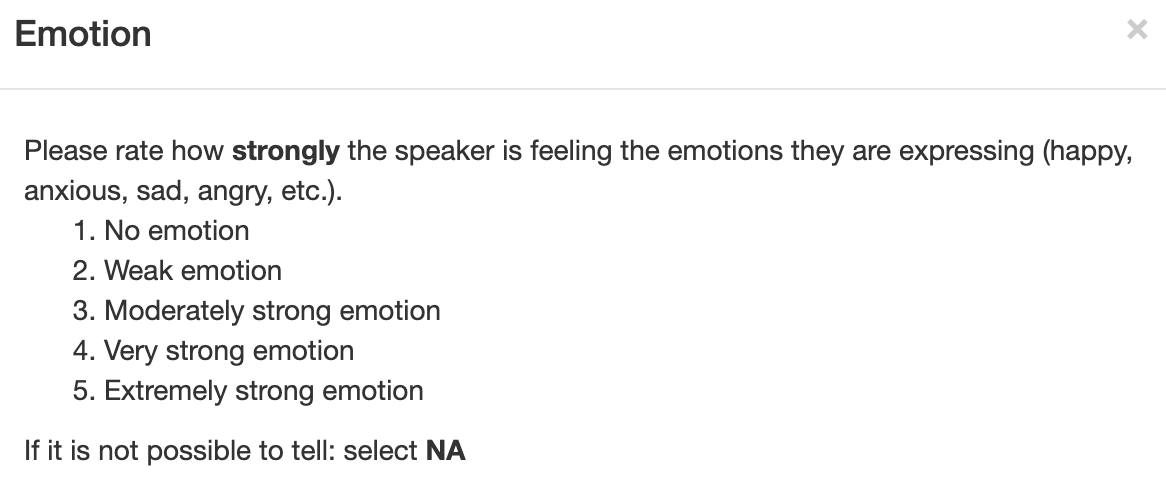}
    \includegraphics[width=8cm]{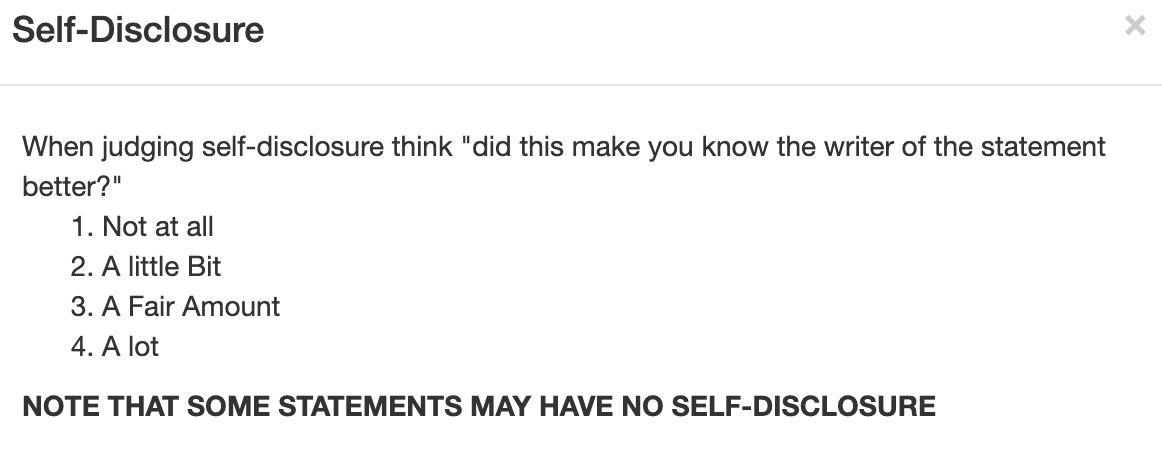}
    \includegraphics[width=8cm]{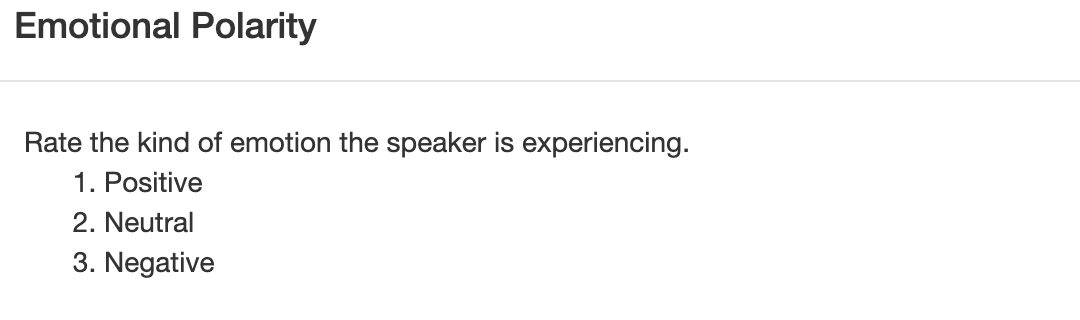}
    \caption{Turn Level Annotation Category and Rating Descriptions}
    \label{fig:turnempathy}
\end{figure}

\begin{figure}[ht!]
    \centering
    \includegraphics[width=8cm]{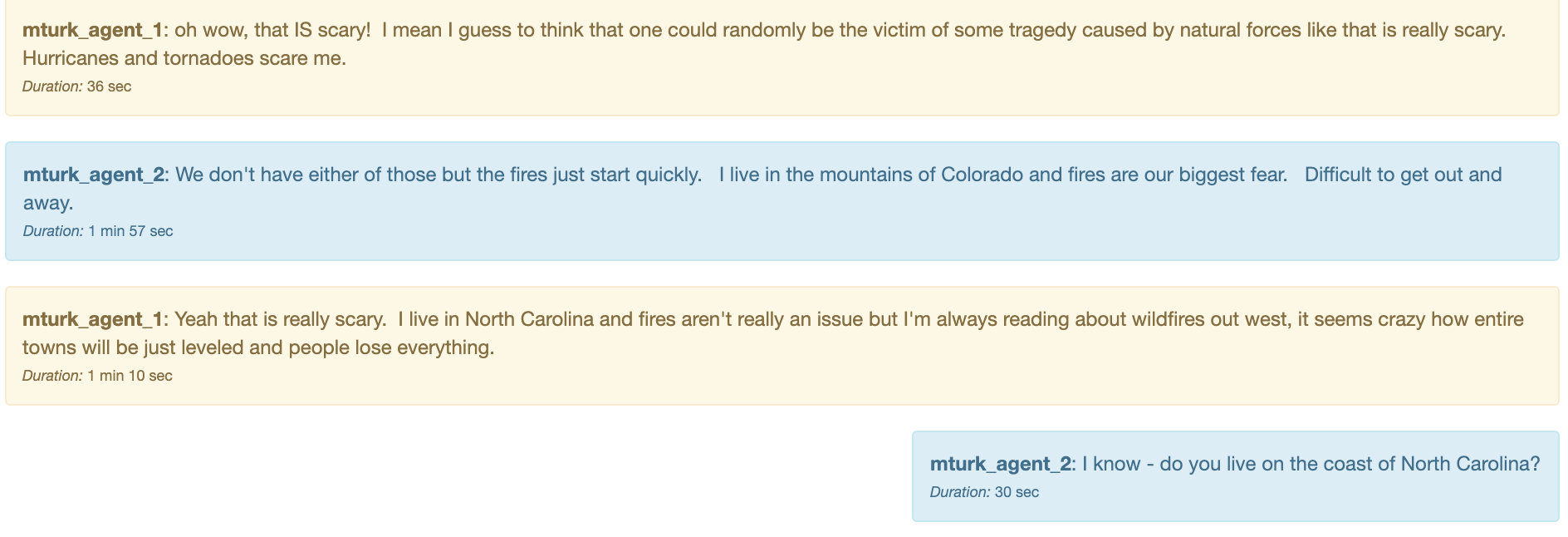}
    \includegraphics[width=8cm]{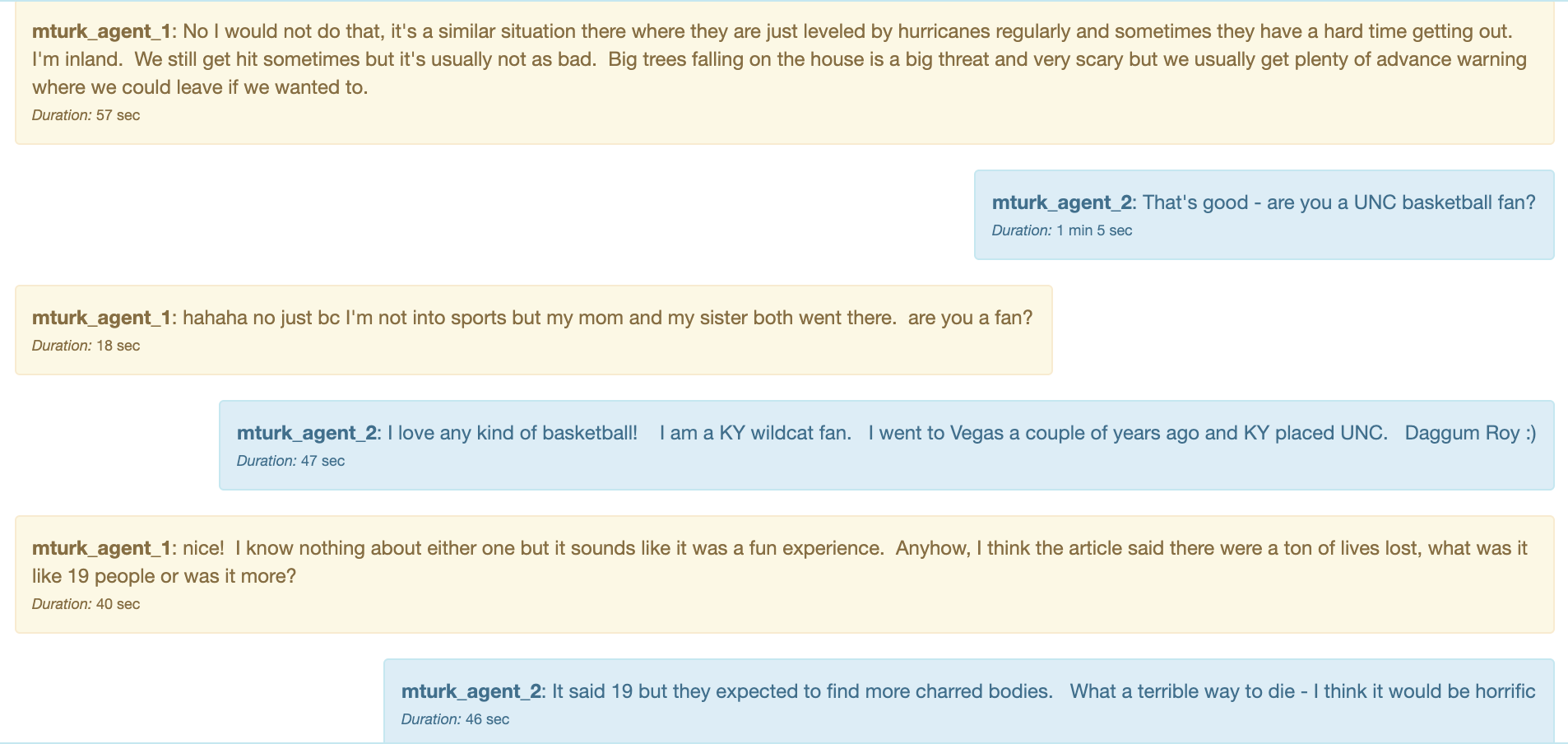}
    \caption{Illustration of the empathetic conversation between two workers about a particular news article.}
    \label{fig:convo}
\end{figure}
\begin{figure*}[ht!]
    \centering
    \includegraphics[width=\linewidth]{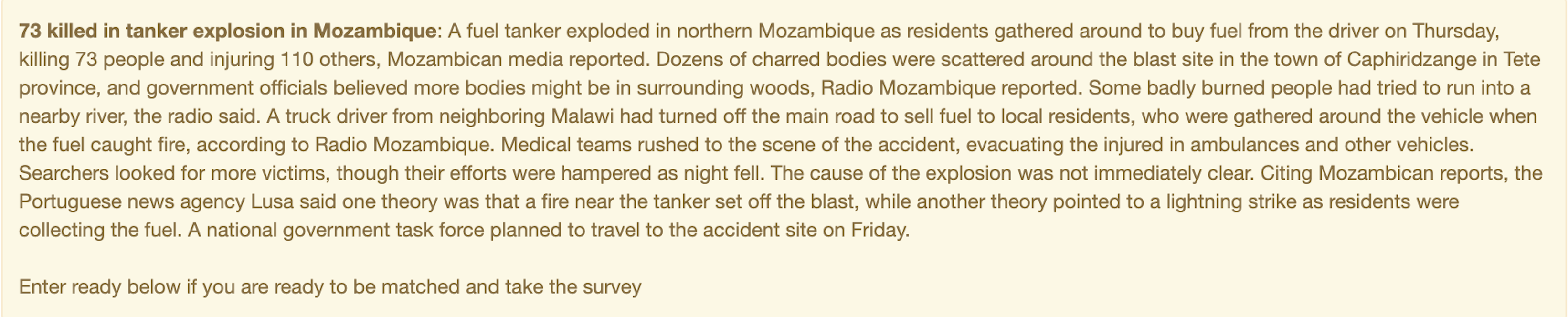}
    \caption{Illustration of a news article provided to workers.}
    \label{fig:conv_article}
\end{figure*}

\begin{figure*}[ht!]
    \centering
    \includegraphics[width=\linewidth]{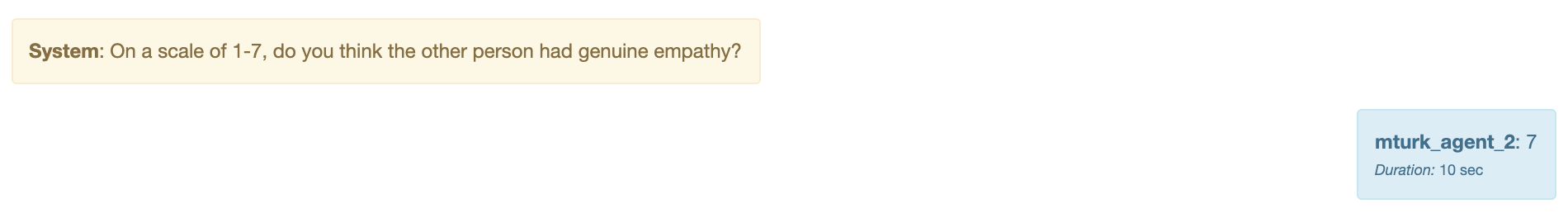}
    \caption{Screenshot of the other-report empathy provided to workers.}
    \label{fig:other_report}
\end{figure*}

\clearpage

\section{Detailed Turn-Level Annotation Breakdown}
\label{sec:turn-iaa-breakdown}
As seen in Figure~\ref{fig:turnempathy}, there are several turn-level annotations by crowd workers. After several pilots we focused on independent tasks with full conversational history after each annotation. This led to the highest inter-annotator agreement. We break down the average per-conversation Krippendorff's alpha in order to understand which aspects are most difficult to annotate. In Table~\ref{tab:turn-level-iaa-by-type} we observe that the empathy IAAs are quite low. This is notable, since annotations are less consistent when there are conversations with extremely low Empathy and Self-Disclosure. Also, by averaging only per-conversation, we are showing a worse IAA than overall which explains why none of the individual annotations score higher than the overall Krippendorff's alpha. Nonetheless, this offers insight into which aspects of annotation are difficult. In an annotation with trained research assistants, we found the same trend.

One common concern about our turn-level annotation is that the Krippendorff's alpha is below 0.667. As \citet{craggs-wood-2005-squibs} suggest,
\begin{quote}
When inferring reliability from agreement, a common error is to believe that there
are a number of thresholds against which agreement scores can be measured in order to
gauge whether or not a coding scheme produces reliable data. Most commonly this
is Krippendorff’s decision criterion, in which scores greater than 0.8 are considered
satisfactory and scores greater than 0.667 allow tentative conclusions to be drawn~\cite{krippendorff2004reliability}. The prevalent use of this criterion despite repeated advice that
it is unlikely to be suitable for all studies~\cite{carletta-1996-assessing,eugenio2004kappa,krippendorff2004reliability} is probably due to a desire for a simple system that can be easily
applied to a scheme. Unfortunately, because of the diversity of both the phenomena
being coded and the applications of the results, it is impossible to prescribe a scale
against which all coding schemes can be judged.    
\end{quote}

\section{Dataset Demographic Statistics}
\label{sec:demo}
There were 44 workers who self-identified as male and 31 who self-identified as female workers who participated in the conversations. We included an option for gender non-disclosure/other; however, this was not selected by participants. 
Tables \ref{table:education} and \ref{table:race} illustrate the distribution of education, and race.
 \begin{table}[ht!]
 \centering
 \begin{tabular}{p{5.5cm}c} 
 \multicolumn{2}{c}{\textbf{Education}} \\
 \hline
  Less than a high school diploma & 0 \\
  \hline
  High School diploma & 8 \\
  \hline
   Technical/Vocational School & 2 \\
  \hline
   Some college but no degree & 13 \\
  \hline
   Two year associate degree & 9 \\
  \hline
   Four year bachelor's degree & 34 \\ 
  \hline
   Postgradute or professional degree & 9 \\
  \hline
 \end{tabular}
 \caption{Distribution of the education level of corpus participants.}
 \label{table:education}
 \end{table}
 \begin{table}[ht!]
 \centering
 \begin{tabular}{p{5.75cm}c} 
 \multicolumn{2}{c}{\textbf{Race/Ethnicity}} \\
 \hline
  White & 56 \\
  \hline
  Hispanic or Latino & 8 \\
  \hline
 Black or African American & 6 \\
  \hline
  Native American or American Indian & 0 \\
  \hline
   Asian/Pacific Islander & 4 \\ 
  \hline
  Other & 1 \\
  \hline
 \end{tabular}
 \caption{Distribution of the race/ethnicity of corpus participants.} 
 \label{table:race}
 \end{table}
\section{Sample of Conversation}
Table \ref{table:4} and Table \ref{table:5} display the differences in the first three turns of two conversations about the same article.
\begin{table*}[ht!]
\centering
\begin{tabular}{cp{32em}} 
\hline
Person 1 & This article just brings home how tough cancer is and how anyone at any time can find themselves having to deal with this terrible illness. \\ 
\hline
Person 2 & My thoughts exactly. I feel that this disease really can impact anybody from any class.
\\ \hline
Person 1 & Exactly! This is something that can't be stopped with money or status or anything like that and brings us all down to the same level.
\\ \hline
Person 2 & I just with there were more clear sets of information regarding cancer. There seems to be too much misinformation.
\\ \hline
Person 1 & There does seem to be a lot of misinformation. But I think it's in part because even the same kind of cancer doesn't act the same way with everyone. Each cancer experience is so different and yet in the end, it's the same disease. I think that is what makes it so frustrating.
\\ \hline
Person 2 & Right but I guess I mean more-so related to basic causes and how a person even gets cancer. It just seems so random and leaves a ton of people fearful that they will get it.
\\ \hline
\end{tabular}
\caption{First three turns of an example conversation where workers had high self-report empathy.}
\label{table:4}
\end{table*}
\begin{table}[ht!]
\centering
\begin{tabular}{cp{15em}} 
\hline
Person 1 & what did you think about this article \\
  \hline
Person 2 & it was interesting \\
  \hline
Person 1 & It was... cancer is such a sad subject  no matter what\\
 \hline
Person 2 & yes it is sad to see in anyone\\
 \hline
Person 1 & I don't get why it has not been cured yet\\
 \hline
Person 2 & probably deliberate \\
 \hline
\end{tabular}
\caption{First three turns of an example conversation where both workers had low self-report empathy.}
\label{table:5}
\end{table}
\section {NN models - training condition}
During training the Bi-RNN model, we use \textit{fasttext} pretrained embedding with 300-dimensions, dropout of 0.2\%, and 3 fully connected layers were used after attention layer. Adam optimizer was used to minimize cross-entropy loss with a learning rate of 0.0001 for the fully connected layer. A batch size of 32 was used with single NVIDIA GTX 1080 Ti. We \textit{fine-tune} RoBERTa-base for 10 epoch using Adam optimizer with the learning rate 0.00001 was used to minimize MSE loss. Each of the experiments was repeated 5 times for RNN models and 3 times on average for RoBERTa models. We report the best model performance.
\clearpage
\section{Datasheet for \textit{Empathic Conversations} Dataset}
\label{sec:appendix}
\citet{gebru2018datasheets} stated that the objective to have datasheet is to clearly describe the process of collection, distribution, and maintenance of a dataset. Thus, below, we provide a datasheet for our \textit{Empathic Conversations} dataset, as structured and suggested by \citet{gebru2018datasheets}. Further, this datasheet could be useful to individuals who participated in this study and to policy makers and advocates. If a question is not applicable to our dataset, the answer is \textit{n.a.}
\subsection{Motivation}
\begin{itemize}
    \item {\textbf{For what purpose was the dataset created? Was there a specific task in mind? Was there a specific gap that needed to be filled? Please provide a description.}\\Empathic Conversations Grounded in News Stories was created be used to better train models to improve their perceived empathy in a conversation.}
    \item {\textbf{Who created the dataset (e.g., which team, research group) and on behalf of which entity (e.g., company, institution, organization)?}\\The data was collected at the University of Pennsylvania.}
    \item {\textbf{Who funded the creation of the dataset? If there is an associated grant, please provide the name of the grantor and the grant name and number}\\ This data was partially funded by Jo\~ao Sedoc's Microsoft Research Dissertation Grant.}
\end{itemize}
\subsection{Composition}
\begin{itemize}
    \item {\textbf{What do the instances that comprise the dataset represent (e.g., documents, photos, people, countries)? Are there multiple types of instances (e.g., movies, users, and ratings; people and interactions between them; nodes and edges)? Please provide a description.}\\Each instance is a breakdown of the conversation had between two workers on Amazon Mechanical Turk over a specific news article. Beyond this, there are personality, demographic, turn-level empathy, self-disclosure, emotion, and dialog acts.}
    \item {\textbf{How many instances are there in total (of each type, if appropriate)?}\\A total of 500 conversations from 100 articles with 5 conversations per article.}
    \item {\textbf{Does the dataset contain all possible instances or is it a sample (not necessarily random) of instances from a larger set? If the dataset is a sample, then what is the larger set? Is the sample representative of the larger set (e.g., geographic coverage)? If so, please describe how this representativeness was validated/verified. If it is not representative of the larger set, please describe why not (e.g., to cover a more diverse range of instances, because instances were withheld or unavailable).}\\n.a.}
    \item {\textbf{What data does each instance consist of? “Raw” data (e.g., unprocessed text or images) or features? In either case, please provide a description.}\\Each instance contains a conversation.}
    \item {\textbf{Is there a label or target associated with each instance? If so, please provide a description.}\\Each conversation is accompanied by the corresponding survey information taken from each participant in the conversation, and the article that the conversation is focused around.}
    \item {\textbf{Is any information missing from individual instances? If so, please provide a description, explaining why this information is missing (e.g., because it was unavailable). This does not include intentionally removed information, but might include, e.g., redacted text.}\\Yes, the Amazon Mechanical Turk worker IDs and IP addresses collected from qualtrics were excluded to preserve the privacy of the workers. Additionally the qualtrics survey timing was excluded. All conversations were reviewed and no private information is present.}
    \item {\textbf{Are relationships between individual instances made explicit (e.g., users’ movie ratings, social network links)? If so, please describe how these relationships are made explicit.}\\One relationship between the instances is that multiple conversations consisted of the same participants, but the participants were not aware of that. Another is that some participants referenced different articles that they already had conversations on.}
    \item {\textbf{Are there recommended data splits (e.g., training, development/validation, testing)? If so, please provide a description of these splits, explaining the rationale behind them.}\\n.a.}
    \item {\textbf{Are there any errors, sources of noise, or redundancies in the dataset? If so, please provide a description.}\\n.a.}
    \item {\textbf{Is the dataset self-contained, or does it link to or otherwise rely on external resources (e.g., websites, tweets, other datasets)? If it links to or relies on external resources, a) are there guarantees that they will exist, and remain constant, over time; b) are there official archival versions
    of the complete dataset (i.e., including the external resources as they existed at the time the dataset was created); c) are there any restrictions (e.g., licenses, fees) associated with any of the external resources that might apply to a future user? Please provide descriptions of all external resources and any restrictions associated with them, as well as links or other access points, as appropriate.}\\n.a.}
    \item {\textbf{Does the dataset contain data that might be considered confidential (e.g., data that is protected by legal privilege or by doctor patient confidentiality, data that includes the content of individuals’ non-public communications)? If so, please provide a description.}\\The dataset does not contain confidential information since all information was scraped from news stories.}
    \item {\textbf{Does the dataset contain data that, if viewed directly, might be offensive, insulting, threatening, or might otherwise cause anxiety? If so, please describe why.}\\The dataset does not contain data that might be offensive, insulting, threatening,  or might otherwise cause anxiety.}
\end{itemize}
\subsection{Collection Process}
\begin{itemize}
    \item {\textbf{How was the data associated with each instance acquired? Was the data directly observable (e.g., raw text, movie ratings), reported by subjects (e.g., survey responses), or indirectly inferred/derived from other data (e.g., part-of-speech tags, model-based guesses for age or language)? If data was reported by subjects or indirectly inferred/derived from other data, was the data validated/verified? If so, please describe how.}\\Conversations and Annotations of the conversations were collected through Amazon Mechanical Turk.}
    \item {\textbf{What mechanisms or procedures were used to collect the data (e.g., hardware apparatus or sensor, manual human curation, software program, software API)? How were these mechanisms or procedures validated?}\\Articles used for the conversations were from Empathic News Reactions (\url{https://github.com/wwbp/empathic_reactions} and \url{https://drive.google.com/file/d/1A-7XiLxqOiibZtyDzTkHejsCtnt55atZ/view}). Crowdworkers were recruited on Amazon mechanical turk. During phase 1 they completed personality tests and entered demographic information. Then afterwards crowdworkers were matched to read the articles, write essays/fill out surveys and chat. After ending the conversation, they rated the other person's empathy. Finally, turn-level annotations of Empathy, Emotion, Self-Disclosure was done via AMT crowdworkers and dialogue acts were labeled by research assistants.}
    \item {\textbf{If the dataset is a sample from a larger set, what was the sampling strategy (e.g., deterministic, probabilistic with specific sampling probabilities)?}\\The 100 articles used were sampled from the  Empathic News Reactions (\url{https://github.com/wwbp/empathic_reactions} and \url{https://drive.google.com/file/d/1A-7XiLxqOiibZtyDzTkHejsCtnt55atZ/view}). The conversations with the highest empathy and distress scores were selected.}
    \item {\textbf{Who was involved in the data collection process (e.g., students, crowdworkers, contractors) and how were they compensated (e.g., how much were crowdworkers paid)?}\\Subsequent demographic information is listed in Tables \ref{table:education} and \ref{table:race}, and Section \ref{taskimp} were performed by crowd workers found through Amazon Mechanical Turk.}
    \item {\textbf{Over what timeframe was the data collected? Does this timeframe match the creation timeframe of the data associated with the instances (e.g., recent crawl of old news articles)? If not, please describe the timeframe in which the data associated with the instances was created.}\\Dataset was collected during the timeframe of June 2019-November 2019 for the conversations and turn-level annotations: May 2020-August 2020 and February 2021-May 2022.}
    \item {\textbf{Were any ethical review processes conducted (e.g., by an institutional review board)? If so, please provide a description of these review processes, including the outcomes, as well as a link or other access point to any supporting documentation.}\\The IRB approved our study. We stated all of the goals of the research and submitted all the surveys and screenshots of the interactions.}
    \item {\textbf{Does the dataset relate to people? If not, you may skip the remainder of the questions in this section.}\\Yes, some news articles were about specific individuals. Also, the conversations are between two people and their personality/demographic information is collected.}
    \item {\textbf{Did you collect the data from the individuals in question directly, or obtain it via third parties or other sources (e.g., websites)?}\\Directly using Amazon Mechanical Turk}
    \item {\textbf{Were the individuals in question notified about the data collection? If so, please describe (or show with screenshots or other information) how notice was provided, and provide a link or other access point to, or otherwise reproduce, the exact language of the notification itself.}\\Yes the participants were consented into the study}
    \item {\textbf{Did the individuals in question consent to the collection and use of their data? If so, please describe (or show with screenshots or other information) how consent was requested and provided, and provide a link or other access point to, or otherwise reproduce, the exact language to which the individuals consented.}\\Participants were informed that their personality, demographic, essays, and conversations will be used for research purposes by people different from us, the collector of these information. The wording we used to communicate with the participants are provide in \ref{ethics} of the main paper.}
    \item {\textbf{Has an analysis of the potential impact of the dataset and its use on data subjects (e.g., a data protection impact analysis)been conducted? If so, please provide a description of this analysis, including the outcomes, as well as a link or other access point to any supporting documentation.}\\Analysis of the potential impact of the dataset is laid out in Section 8.}
\end{itemize}
\subsection{Preprocessing/cleaning/labeling}
\begin{itemize}
    \item {\textbf{Was any preprocessing/cleaning/labeling of the data done (e.g., discretization or bucketing, tokenization, part-of-speech tagging, SIFT feature extraction, removal of instances, processing of missing values)? If so, please provide a description. If not, you may skip the remainder of the questions in this section.}\\The following step were taken to process the data: 1) Gathering New Articles: News Articles were selected from the Empathy New Reactions dataset\footnote{\url{https://github.com/wwbp/empathic\_reactions} and articles \url{https://drive.google.com/file/d/1A-7XiLxqOiibZtyDzTkHejsCtnt55atZ/view}} with 100 articles with the highest empathy and distress ratings, 2) Labeling for the Turn Level Annotation collection: Empathy, Emotion, Self-Disclosure, and Emotional Polarity were the annotation categories for the turns in each conversation.}
    \item {\textbf{Was the “raw” data saved in addition to the preprocessed/cleaned/labeled data (e.g., to support unanticipated future uses)? If so, please provide a link or other access point to the “raw” data.}\\The raw unprocessed data (consisting of conversations) is saved on a secure server as per the IRB requirements.}
    \item {\textbf{Is the software used to preprocess/clean/label the instances available? If so, please provide a link or other access point.}\\This is available by request.}
\end{itemize}
\subsection{Uses}
\begin{itemize}
    \item {\textbf{Has the dataset been used for any tasks already? If so, please provide a description. }\\This dataset is used to model empathy and distress and to \textit{fine-tune} a conversation empathic model; results are provided in this article. The dataset has also been used to create a empathic chatbot.}
    \item {\textbf{Is there a repository that links to any or all papers or systems that use the dataset? If so, please provide a link or other access point.}\\Yes, available by request.}
    \item {\textbf{What (other) tasks could the dataset be used for?}\\The author's suggest that this dataset be used to enhance tasks in which empathy is a module and this dataset would be distributed for research purposes.}
    \item {\textbf{Is there anything about the composition of the dataset or the way it was collected and preprocessed/cleaned/labeled that might impact future uses? For example, is there anything that a future user might need to know to avoid uses that could result in unfair treatment of individuals or groups (e.g., stereotyping, quality of service issues) or other undesirable harms (e.g., financial harms, legal risks) If so, please provide a description. Is there anything a future user could do to mitigate these undesirable harms?}\\ The demographic information of the participants in the conversations is known, so demographic fairness checks can be used.}
    \item {\textbf{Are there tasks for which the dataset should not be used? If so, please provide a description.}\\The creation of malicious bots.}
\end{itemize}
\subsection{Distribution}
\begin{itemize}
    \item {\textbf{How will the dataset will be distributed (e.g., tarball on website, API, GitHub)? Does the dataset have a digital object identifier (DOI)?}\\The dataset will be distributed via requests and for research purposes. }
    \item {\textbf{When will the dataset be distributed?}\\Via requests and for research purposes.}
    \item {\textbf{Will the dataset be distributed under a copyright or other intellectual property (IP) license, and/or under applicable terms of use (ToU)? If so, please describe this license and/or ToU, and provide a link or other access point to, or otherwise reproduce, any relevant licensing terms or ToU, as well as any fees associated with these restrictions.}\\The corpus will be available under an Academic License (free of cost). Researchers will download a pdf and will supply their intended use. The process will require the institution to sign the agreement. The license will be a standard FDP data transfer form. It should be easy for institutions to obtain a signature.}
    \item {\textbf{Have any third parties imposed IP-based or other restrictions on the data associated with the instances? If so, please describe these restrictions, and provide a link or other access point to, or otherwise reproduce, any relevant licensing terms, as well as any fees associated with these restrictions.}\\Third parties may not redistribute the data.}
    \item {\textbf{Do any export controls or other regulatory restrictions apply to the dataset or to individual instances? If so, please describe these restrictions, and provide a link or other access point to, or otherwise reproduce, any supporting documentation.}\\n.a.}
\end{itemize}
\subsection{Maintenance}
\begin{itemize}
    \item {\textbf{Who is supporting/hosting/maintaining the dataset?}\\The authors of the paper provide maintenance and support.}
    \item {\textbf{How can the owner/curator/manager of the dataset be contacted (e.g., email address)? }\\By email jsedoc@stern.nyu.edu.}
    \item {\textbf{Is there an erratum? If so, please provide a link or other access point.}\\n.a.}
    \item {\textbf{Will the dataset be updated (e.g., to correct labeling errors, add new instances, delete instances)? If so, please describe how often, by whom, and how updates will be communicated to users (e.g., mailing list, GitHub)?}\\The authors keep a mailing list of the users of this dataset and all changes will be communicated by the authors of this paper.}
    \item {\textbf{If the dataset relates to people, are there applicable limits on the retention of the data associated with the instances (e.g., were individuals in question told that their data would be retained for a fixed period of time and then deleted)? If so, please describe these limits and explain how they will be enforced.}\\n.a.}
    \item {\textbf{Will older versions of the dataset continue to be supported/hosted/maintained? If so, please describe how. If not, please describe how its obsolescence will be communicated to users.}\\Authors provide support for all versions that are in use.}
    \item {\textbf{If others want to extend/augment/build on/contribute to the dataset, is there a mechanism for them to do so? If so, please provide a description. Will these contributions be validated/verified? If so, please describe how. If not, why not? Is there a process for communicating/distributing these contributions to other users? If so, please provide a description.}\\From the ethical perspective the same protocol that is developed for this work should be used in addition to any new code of ethics if applicable. Documents are available and can be distributed to the eligible people who want to extend the work. }
\end{itemize}

\end{document}